# A repeated-measures study on emotional responses after a year in the pandemic

Maximilian Mozes[1,2,3], Isabelle van der Vegt[1], Bennett Kleinberg[4,1] *

*All authors contributed equally to this paper*

[1] Department of Security and Crime Science, University College London, UK
[2] Dawes Centre for Future Crime, University College London, UK
[3] Department of Computer Science, University College London, UK
[4] Department of Methodology and Statistics, Tilburg University, The Netherlands

* Corresponding author: Bennett Kleinberg, <u>bennett.kleinberg@tilburguniversity.edu</u>

**Abstract**
The introduction of COVID-19 lockdown measures and an outlook on return to normality are demanding societal changes. Among the most pressing questions is how individuals adjust to the pandemic. This paper examines the emotional responses to the pandemic in a repeated-measures design. Data (*n*=1698) were collected in April 2020 (during strict lockdown measures) and in April 2021 (when vaccination programmes gained traction). We asked participants to report their emotions and express these in text data. Statistical tests revealed an average trend towards better adjustment to the pandemic. However, clustering analyses suggested a more complex heterogeneous pattern with a well-coping and a resigning subgroup of participants. Linguistic computational analyses uncovered that topics and *n*-gram frequencies shifted towards attention to the vaccination programme and away from general worrying. Implications for public mental health efforts in identifying people at heightened risk are discussed. The dataset is made publicly available.

**Introduction**
The COVID-19 pandemic has been dominating people's lives for far more than a year now. Despite the success of ongoing vaccination programmes in various countries, lockdown and social distancing measures remain in place around the world. In the United Kingdom, however, at the time of writing this paper, society is increasingly opening up, and residents can visit shops, bars and restaurants and cultural and entertainment venues (e.g., museums, sports stadiums and cinemas) are welcoming visitors again [1].

The impact of the pandemic on people's lives has been studied widely by researchers in the last year, and the natural language processing (NLP) research community has focussed on linguistic phenomena related to the pandemic in a range of studies [2–4]. In April 2020, shortly after the start of the SARS-CoV-2 outbreak and the COVID-19 pandemic, the COVID-19 *Real World Worry Dataset* (RWWD) [5] was presented - a dataset of 5,000 long and short texts in which participants expressed their feelings about the pandemic. The text data were accompanied by 9-point-scale ratings for a set of nine emotional variables (e.g., anger, anxiety, happiness, sadness and worry) and aids in understanding the impact of the pandemic on people's mental wellbeing and the nexus between emotion and text expression.

The study found that participants reported predominantly negative emotions such as anxiety, worry, sadness and fear. Furthermore, using the text data to understand the participants' concerns and emotional responses, topic modelling methods uncovered that people were mainly writing about worries related to family, friends, employment, and the economy. A different study further showed that the text data reveal gender differences in the concerns and worries of people [6]. The dataset has since been used to understand emotional responses to COVID-19 through the lens of text data (e.g., [7]).

The original RWWD dataset was collected when the UK was hit hard by COVID-19, and residents were instructed to stay at home and leave their houses only for essential purposes. Large parts of the economy, including the hospitality and retail sectors, were completely locked down at that time [8]. In the current paper, we follow up on the work by Kleinberg et al. [5], presenting a new corpus of text data related to the COVID-19 pandemic, written by the same participants one year after the original data collection. We asked the participants to again write about their feelings concerning the pandemic and provide the same emotional measures. Using this unique within-subjects corpus, we assess to what extent people's emotions changed as society is slowly opening up and ongoing vaccination programmes are paving the way to an eventual return to normality. Specifically, we conduct statistical comparisons between emotion scores from both phases and assess individual differences based on topic modelling approaches and word frequency comparisons between texts of both data collection phases. Using clustering analyses, we also test for potential subgroups of participants with respect to emotion change.

Our findings show that, compared to the first phase, the participants demonstrate stronger positive emotional responses in the second phase (e.g., happiness and relaxation) and score lower on negative ones (e.g., fear and anxiety). Data clustering algorithms, however, suggest that this pattern is far from homogenous. Two cluster groups are identified based on the emotion change scores, a well-coping and a resignation cluster. *n*-gram frequency analysis and topic modelling indicate that text data reveal a focus shift from the first to the second phase.

*Text-based analyses around COVID-19*
Since the emergence of COVID-19, numerous studies have examined the effect of the pandemic. Various behavioural research efforts have used text-based analyses concerning COVID-19 [9–11]. A large number of studies made use of Twitter data to perform textual analyses. Examples include the spread of misinformation on Twitter [12,13], emotional consequences of lockdowns studied through Twitter (and Weibo) data [14], as well as the association between US partisanship and sentiment towards COVID-19 measures [15]. Other text data sources include Reddit, which has, for example, been used to characterise the linguistic difference between posters on the r/Coronavirus and the r/China_flu page [16]. Based on word frequency measures, the language difference between the two subreddits was low in January and February of 2020, but diverged for the rest of the year, indicating the emergence of two distinct communities characterised by their attitude towards the virus. In another study of Reddit data, linguistic features including the Linguistic Inquiry and Word Count software (LIWC) [17], sentiment analysis, and word frequencies were used to study changes in mental health discourse across 15 subreddits [18]. All in all, the studies above demonstrate how the availability of text data on social media platforms enables researchers to study the effects of COVID-19 on a large scale.

Previous research has also specifically examined emotional responses to the COVID-19 crisis. Using 122M tweets posted throughout 2020, [19] propose that the psycho-social response to the pandemic occurs in three phases. Following Strong's 1990 model of the sociological reaction to fatal epidemics, refusal is followed by anger, which in turn is followed by acceptance. The authors made use of keywords from the original Strong (1990) paper, the LIWC, and lexicons for emotions, moral foundations, and pro-social behaviour to examine the occurrence of the three phases in the Twitter data. Aiello et al. [19] note that refusal and acceptance decreased throughout 2020, whereas acceptance increased. When new waves of COVID-cases emerged, new waves of anger were observed in the language use on Twitter. In a similar endeavour, Stella et al. [20] examined social discourse within tweets from Italy straight after the first lockdown in March 2020. Tweets were collected by searching three hashtags related to the lockdown, after which networks of hashtags were constructed by examining co-occurrences of hashtags in tweets. The tweets were subsequently analysed for their sentiment polarity using the NRC lexicon. By exploring the emotional valence in the identified hashtag networks, the authors note that they observed co-existence of anger and fear with trust, solidarity and hope in the corpus. Emotional responses to the pandemic have also been examined by other means than text data, for instance in an exploration of the emotional appreciation of humour (memes, jokes) related to COVID-19 [21]. In an online survey in April 2020, different clusters of emotional responses were identified (e.g., optimists and pessimists) which were found to be related to several variables such as gender, pre-existing health conditions, and political orientation [22]. In related research using an online task, participants who scored high on COVID-19 fear and negative emotion also judged words relating to COVID-19 more negatively [23]

These studies are united by using text data as a proxy for real-life behaviour or attitudes. Among the many potentials of text data is the possibility to use observational data - often produced as a by-product of human online behaviour (e.g., participation in online forums, see [24]) - as the lens through which we study human emotions, attitudes and behaviour. At the same time, pure observational text data often lack *ground truth* in the form of known emotional states, attitudes or even behaviours. The current study addresses the ground truth problem by collecting primary data in a one-year lagged within-subjects design, thereby extending the research questions.

*Aims of this paper*
This paper makes two key contributions. First, we present and make available a within-subjects follow-up dataset for the RWWD, comprising people's emotional responses to the COVID-19 pandemic and corresponding text data at two moments in time. Doing so helps us study the phenomenon in a longitudinal sense and may allow us to pose new research questions beyond observational snapshot studies. Second, this paper seeks to shed further light on the emotional responses to the Coronavirus pandemic. Specifically, we are interested in the change that happened and consequently may have led to different emotional responses between the onset of the pandemic and a year later when the UK was about to open the country again. We aim to answer two questions: (1) How did people's emotional responses change one year after the pandemic? (2) How can text data help us understand the change in emotional responses?

By connecting textual responses to emotional states, we specifically aim to explore how emotions are reflected in human language use. Such an investigation might help us exploit potential relationships between these two modalities, and to obtain insights into emotional states solely based on human language data. Our findings could provide helpful insights for health professionals and policy makers in tackling mental health issues as a consequence of the COVID-19 pandemic.

**Method**

*Data availability statement*
The data collected for and used in this paper are publicly available at https://osf.io/szxda/?view_only=038eaa81edfa4026af3f6f049415ef39

*Ethics statement*
The study has been reviewed and approved by the IRB of University College London. All participants were informed about the purpose and procedure of the study and provided informed consent and were over the age of 18. All methods were performed in accordance with the guidelines of the Declaration of Helsinki.

*Data collection*
Our data consisted of repeated measures of the same sample at two different points in time (phase 1 and phase 2, respectively). Data from phase 1 were collected in April 2020 [5] and consisted of $n_1$ = 2491 participants who each reported their emotions about the Coronavirus pandemic and expressed in text form how they felt. We contacted each participant who contributed to the first phase via the crowdsourcing platform Prolific Academic and notified them of a new data collection precisely one year after the original study. We used the identical questionnaire, task interface (with Qualtrics software), participant payment and exclusion criteria. The second phase of data collection ran from March to April 2021.

The initial sample of phase 2 consisted of $n_2$ = 1839. After applying the following exclusion criteria, we retained $n$=1716 participants: participants who did not write an English text ($n$=67), wrote more than 20% punctuation ($n$=6), did not input a valid Prolific id ($n$=33, e.g., due to a typo) or were duplicates in the data ($n$=17) were excluded. We then merged the two data collection phases: $n$=18 participants who completed phase 2 were not in the final dataset from phase 1 (we reached out to all participants who participated in phase 1 – but did not necessarily complete the task or end up in the final dataset, e.g. due to timing out in the data collection task of phase 1 or due to being excluded from the dataset in phase 1 due to more than 20% punctuation). The full final sample consisted of $n$=1698 participants who provided complete data for both data collection phases. Of that sample, 67.37% were female, 31.33% male, and 1.30% did not provide gender information. The mean age was 36.22 years ($SD$=11.66) with a range from 18 to 83 (an outlier with an age of 961 was excluded from the age calculation but retained for all other analyses). All participants were UK residents with 88.40% being born in the UK. The majority were in full-time (55.83%) or part-time employment (22.08%). 10.54% of the sample were not in paid work (e.g., retired or homemaker) and 7.53% were unemployed. Being an active student (regardless of employment) applied to 17.90% of the sample.

*Context of data collection*

Data for phase 1 of this study were collected during the beginning of lockdown in the UK (April 2020), where death tolls attributable to the virus were steadily increasing. At that time, Queen Elizabeth II has just addressed the nation via a television broadcast, and Prime Minister Boris Johnson was admitted to intensive care in a hospital for COVID-19 symptoms [25]. One year later, in phase 2 of this study (March-April 2021), many UK residents had been vaccinated, and schools, retail and the hospitality sectors were (partially) re-opened. Reports of the newly identified delta-variant of the Coronavirus just started to emerge at this time.

*Emotion data*

Each participant was asked how they felt about the Coronavirus pandemic at the moment of data collection for each of eight emotions and emotional states (anger, anxiety, desire, disgust, fear, happiness, relaxation, sadness [26]; and worry. They indicated their emotions for each on a 9-point Likert scale (1=very low, 5=moderate, 9=very high). In addition, they were asked to pick which of the eight emotions best described their feeling if they had to choose just one.

*Text data*

After the participants self-reported their emotions, we asked them to express how they felt at that moment about the COVID-19 situation in text form. Their task was to write one long text of at least 500 characters and a shorter one that expresses their feeling in Tweet length (max. 280 characters). Table 1 shows verbatim excerpts of one participant for both phases and text lengths.

Table 1. *Excerpts from both data collection phases (long and short texts)*

|  | **Phase 1 (April 2020)** | **Phase 2 (April 2021)** |
| --- | --- | --- |
| Long | *Fearful and anxious.*<br>*Nervous stomach and racing thoughts.*<br>*Worry for the world and what it will become.*<br>*Scared for my family and my friends.*<br>*Sad for the elderly and those who are totally alone.*<br>*Sick of trying to get food into the house!*<br>*Feels like a survival game or a science fiction novel.*<br>*Unable to concentrate on anything for m[ore] than a couple of minutes - brain like a moth around a light.*<br>*Cross with people who are flouting the rules.*<br>*Worry that this is a virus we know nothing about and so can't possibly know the outcome 6 months, a year or even 2 years down the track.*<br>*Every time I go shopping I see germs and fear other people - I see the fear in them too.*<br>*Want to wak[e] up and it has all gone away* | *Fearful about the easing of lockdown in case cases rise and we have to lockdown again.*<br>*Still worried that someone I love will get sick but trying to keep the odds in perspective.*<br>*Hoping that the government stick to their promise of 'data not dates'.*<br>*Wondering when it will all be over and we can live normal lives again.*<br>*Sick of hearing about the vaccine. Will not be having it but wonder how that will affect my freedom.*<br>*Angry about the possibility of 'immunity passports'.*<br>*Worried that masks will become the norm - hate faceless people.* |
| Short | *Coronavirus can just do one. Had enough now. #stayhome #covidiots or this will go on much longer than we could ever have envisaged.* | *Sick to the back teeth of Covid, lockdowns, masks and vaccines.*<br>*Wake me up when it's over.* |

*Additional variables*

We also asked participants how well they thought they were able to express their feelings in text (in general, the short text and in the long text) and how often they use Twitter (specifically: being on Twitter, responding to Tweets and tweeting themselves). Each variable was measured on a 9-point Likert scale (these data are available in the full dataset and are not analysed here). A comprehensive range of sociodemographic variables is collected by default through the crowdsourcing platform Prolific but not further examined here yet available in the shared dataset.

*Analysis plan*

We assess our key research questions as follows. To test how emotional responses changed, we conduct within-subjects *t*-tests and report the corresponding Cohen's *d* effect size with confidence intervals and the Bayes Factor. The *d* effect size expresses the absolute magnitude of an effect with values of 0.2, 0.5 and 0.8 representing small, moderate and large effects, respectively [27]. The Bayes Factor (BF) is derived from the Bayesian within-subjects *t*-test [28] by testing how likely the data are under two competing hypotheses (here: the null hypothesis of no difference vs a two-sided alternative hypothesis). We used the *bayesfactor* R package [29] with default, non-informative settings for the prior distribution (i.e., a Cauchy prior on the effect size with a scale parameter of $\sqrt{2}/2$, for details see [30]. That prior was chosen because of its desirable properties for the current context (i.e., changes in emotional responses within-subjects after a year in a pandemic), which include a high prior probability density around zero, a symmetrical spread of the prior probability to positive and negative effect sizes (i.e., reflecting the two-sided alternative hypothesis), and its "small influence on the posterior distribution, such that most of the diagnosticity comes from the likelihood of the data" [30]. A *BF*=1.00 implies that both are equally probable. BFs larger than 10 and 100 can be regarded as strong and extreme evidence in favour of the alternative hypothesis, respectively [31]. We also report the Pearson correlation between the two measurements.

In addition to the statistical comparisons, we furthermore examined whether the change in emotions is homogenous across the sample. We do this with *k*-means clustering.

The text data are analysed in three different ways, each allowing us to understand the content in a different manner. *i)* We used the LIWC to measure differences between phase 1 and phase 2 in psycholinguistic categories (e.g., personal concerns, time references). The LIWC employs a dictionary-based approach that counts how many predefined words occur in each text. All LIWC categories were tested in a data-driven exploratory manner. *ii)* Structural topic modelling was used to identify co-occurrences of words that may form latent topics. We tested how topics changed between the two data collection moments. *iii)* We used *n*-gram frequency analysis to test which (sequences of) words differentiated the most between phase 1 and phase 2. That approach may inform us about how foci of the texts shifted.

**Results**

*Change in emotions*

Table 2 shows that there is a substantial decrease in anxiety (*d*=0.55 [0.48; 0.62]), fear (*d*=0.75 [0.68; 0.82]) and worry (*d*=0.86 [0.78; 0.93]). There were small-to-moderate increases in desire (*d*=0.30 0.24; 0.36]), happiness (*d*=0.32 [0.26; 0.38]) and relaxation (*d*=0.26 [0.19; 0.32]). Marginal-to-small effects were observed for a decrease in anger (*d*=0.10 [0.04; 0.17]) and disgust (*d*=0.10 [0.04; 0.17]). Overall, these findings suggest that the participants experienced considerably fewer negative emotions and somewhat more positive emotions. Anger and disgust remained rather stable. The correlations between the phase 1 and phase 2 scores ranged from $r$=0.34 to $r$=0.50.

Table 2. *Mean (SD) emotion scores in both phases with effect size, BF and correlation for paired comparison*

| Emotion | Phase 1 | Phase 2 | Correlation $r$ [99% CI] | Difference | Cohen's $d$ [99% CI] | Bayes Factor |
|---|---|---|---|---|---|---|
| *anger* | 3.81 (2.21) | 3.55 (2.37) | 0.40 [0.35; 0.46] | -0.26 (2.51) | 0.10 [0.04; 0.17] | 213.89 |
| *anxiety* | 6.46 (2.31) | 5.08 (2.51) | 0.46 [0.40; 0.50] | -1.38 (2.52) | 0.55 [0.48; 0.62] | > $10^{95}$ |
| *desire* | 2.86 (1.98) | 3.57 (2.26) | 0.37 [0.32; 0.42] | 0.71 (2.38) | 0.30 [0.24; 0.36] | > $10^{30}$ |
| *disgust* | 3.12 (2.12) | 2.87 (2.18) | 0.36 [0.31; 0.42] | -0.25 (2.42) | 0.10 [0.04; 0.17] | 219.97 |
| *fear* | 5.60 (2.28) | 3.77 (2.28) | 0.43 [0.38; 0.48] | -1.83 (2.44) | 0.75 [0.68; 0.82] | > $10^{163}$ |
| *happiness* | 3.60 (1.87) | 4.32 (2.01) | 0.34 [0.28; 0.39] | 0.72 (2.24) | 0.32 [0.26; 0.38] | > $10^{34}$ |
| *relaxation* | 3.89 (2.11) | 4.54 (2.24) | 0.34 [0.28; 0.39] | 0.65 (2.50) | 0.26 [0.19; 0.32] | > $10^{22}$ |
| *sadness* | 5.55 (2.33) | 4.69 (2.57) | 0.38 [0.33; 0.43] | -0.87 (2.73) | 0.32 [0.25; 0.38] | > $10^{33}$ |
| *worry* | 6.57 (1.75) | 4.95 (2.01) | 0.50 [0.45; 0.54] | -1.62 (1.90) | 0.86 [0.78; 0.93] | > $10^{200}$ |

*Note:* Positive difference scores imply that the values were larger in phase 1 than in phase 2.

*Clustering of emotion change*

The findings from Table 2 show the aggregated mean change for the whole sample, but this may obscure individual differences. To examine whether potential subgroups exist within the sample, we applied *k*-means clustering on the nine emotion change scores (phase 2 - phase 1).

The *k*-means method used Euclidean distance on the nine-dimensional vector representations corresponding to the emotion scores to find the optimal allocation of data points (here: *participants)* into *k* clusters [32]. To determine *k* for the final model, we ran the algorithm for $k \in \{1, ..., 20\}$ and used the within-cluster sum of squares (the "elbow" method) and the silhouette coefficients [33]. Both methods indicated that a cluster model with *k*=2 cluster centres is most suitable to describe the data. The means of each of the emotion change variables per cluster (Table 3) show two distinct groups of participants. We tested for each cluster mean whether it deviated from zero (i.e., no change). All except for anxiety (cluster 2) differed significantly from 0 at *p* < .01.

Cluster 1 consisted of participants who reported marked decreases in anger, anxiety, disgust, fear, sadness and worry, and increases in desire, happiness and relaxation. In contrast, participants in cluster 2 reported higher anger, desire, disgust and sadness, and a decrease in fear, happiness, relaxation and worry.

These patterns lead us to interpret participants in cluster 1 (43.58% of the participants) as well-coping individuals who fared significantly better one year after the start of the pandemic: all negative emotional states (anger, anxiety, disgust, sadness and worry) decreased while all positive ones (desire, happiness, relaxation) increased. The second cluster (56.42% of the participants) showed a more complex pattern: participants were less afraid and less worried but also less happy and less relaxed. Instead, the increases in anger, disgust, sadness and desire, suggest some resignation after one year in the pandemic. Table 4 shows verbatim excerpts of texts of two participants representative for each cluster at the different data collection moments.

Participants in the resignation cluster were marginally younger than those in the well-coping cluster, *d* = 0.16 [99% CI: 0.04; 0.29]. There was no difference in gender between the two clusters, $X^2(1) = 1.86$, p = .173 (we excluded participants from the gender analysis who did not provide gender information). Additional exploratory analyses on predicting cluster membership from the text data using machine learning can be found in the Appendix.

Table 3. *Cluster means (SD) and effect sizes for difference from zero after k-means clustering with the final model of k=2.*

| Emotion variable | Cluster 1 (well-coping) | | Cluster 2 (resignation) | |
|---|---|---|---|---|
| | *M (SD)* | Cohen's *d* [99% CI] | *M (SD)* | Cohen's *d* [99% CI] |
| Anger change | -1.74 (2.08) | 0.83 [0.72; 0.94] | 0.88 (2.19) | 0.40 [0.32; 0.49] |
| Anxiety change | -3.06 (2.13) | 1.44 [1.30; 1.57] | -0.09 (1.96)[ns] | -0.04 [-0.13; 0.04] |
| Desire change | 1.21 (2.44) | 0.50 [0.40; 0.60] | 0.33 (2.27) | 0.14 [0.06; 0.23] |
| Disgust change | -1.36 (2.14) | 0.64 [0.53; 0.74] | 0.61 (2.28) | 0.27 [0.18; 0.35] |
| Fear change | -3.52 (1.94) | 1.81 [1.65; 1.96] | -0.53 (1.94) | 0.28 [0.19; 0.36] |
| Happiness change | 1.88 (2.10) | 0.90 [0.79; 1.01] | -0.19 (1.90) | 0.10 [0.01; 0.18] |
| Relaxation change | 2.16 (2.24) | 0.96 [0.85; 1.08] | -0.53 (2.01) | 0.26 [0.18; 0.35] |
| Sadness change | -2.70 (2.37) | 1.14 [1.02; 1.26] | 0.55 (2.07) | 0.27 [0.18; 0.35] |
| Worry change | -2.48 (1.85) | 1.34 [1.21; 1.47] | -0.97 (1.66) | 0.58 [0.49; 0.67] |
| | | | | |

| Prevalence of cluster | 43.58% (n=740) | 56.42% (n=958) |
|---|---|---|
| Age of cluster members | 37.30 (12.08) | 35.39 (11.27) |
| % female | 66.70% | 69.96% |

*Note*: Negative *change* values indicate that the value of the emotion went down in phase 2 compared to phase 1. Vice versa for positive values. Values of zero imply no change. $^{ns}$ = not significantly different from 0 at p < .01.

Table 4. *Examples of texts written by participants in each cluster (long texts only).*

|  | **Phase 1 (April 2020)** | **Phase 2 (April 2021)** |
|---|---|---|
| Cluster 1 (well-coping) | I am fearful for the health of two relatives; one my husband who has many health issues and I suspect would not survive if he were infected by COVID19; also my daughter in law, a young mother of two [...]. We are self isolating [...] and feel relatively safe at the moment, but until a vaccine is available, the threat will remain. We expect scientists to achieve this within a year. I'm so glad we have the internet and phone to keep in touch with family, and would feel distraught if these lines of communication were cut. [...] | I and my husband have had the first vaccination and it was a joyous moment. Those waiting with us at the venue were eager and happy too. Today our daughter was invited for her jab tomorrow and that makes us feel relief. Our son is a teacher and he will be vaccinated on Tuesday [...]. In the wider world, I fear the re-opening of schools next week may allow cases to rise and I am nervous of that. But overall I feel safer and look forward tentatively to some resumption of normal life. |
| Cluster 2 (resignation) | I feel mostly frustrated not knowing how much longer there is to wait until we are able to go outside, see friends and attend gigs and festivals. I feel lost and more alone than I usually do. I feel confused because I do not know how worried I should be about the virus itself. I am concerned about my own wellbeing and mental health. I feel relaxed because my daily life which I was already feeling aimless in, has stopped and I can enjoy time at home without the weight of pressure I usually feel. [...]. | I feel completely deflated, the government keeping making so many promises they can't keep, prioritising the wrong things and whilst everyone I see is hopeful for things to be more normal by june 21st, our track record has shown me not to trust a single thing, it just feels like there's still too far to go before I will feel any better about the situation, I am severely depressed, I've been unable to work for a year [...] i'm not receiving furlough because of the governments incompetence so I'm a sitting duck waiting for my job to be able to go ahead |

*Linguistic analysis*

Table 5 shows the corpus descriptive statistics for both phases. There were no marked changes between the corpora of the two phases. The linguistic analyses were conducted with the *quanteda* R package [34]. For the subsequent analyses, we use the long texts only. All results for the short texts are available in the Appendix.

Table 5. *Corpus descriptives for both phases (long and short texts).*

| **Variable** | **Phase 1** | **Phase 2** |
|---|---|---|
| Number of tokens (long) | 126.95 (38.01) | 125.39 (31.65) |
| Number of tokens (short) | 27.08 (15.69) | 24.70 (15.64) |
| Number of sentences (long) | 5.70 (2.47) | 5.47 (2.28) |
| Number of sentences (short) | 1.89 (1.06) | 1.72 (0.99) |
| TTR (long) | 0.66 (0.06) | 0.66 (0.06) |
| TTR (short) | 0.89 (0.09) | 0.90 (0.09) |

*Note:* TTR denotes type-token-ratio.

Psycholinguistic variables

In order to capture possible psychological changes reflected in language use, we examined the change in LIWC2015 categories [17] between phase one and phase two. A large number of LIWC categories differed between the two phases for long texts. Table 6 lists the ten categories with the largest BF: the categories "time", "focus on the past", "Tone" (i.e., emotional tone, where a higher value represents more positive emotion), and "relativity" increased in phase two. References to anxiety, negative emotion, the home, as well as pronouns, social words, and interrogators decreased in phase two.

Table 6. *The 10 most changed LIWC categories in long texts (ranked by BF)*

| Category | BF | Cohen's *d* [99% CI] | Wave 2 vs. wave 1 |
|---|---|---|---|
| time | 260.15 | 0.60 [0.54; 0.67] | + |
| anxiety | 257.94 | 0.60 [0.53; 0.67] | - |
| negative emotion | 190.06 | 0.51 [0.44; 0.57] | - |
| focus on the past | 123.55 | 0.40 [0.34; 0.47] | + |
| home | 116.46 | 0.39 [0.33; 0.45] | - |
| emotional tone | 115.95 | 0.39 [0.32; 0.45] | + |
| relativity | 108.74 | 0.38 [0.31; 0.44] | + |
| pronouns | 78.92 | 0.32 [0.26; 0.38] | - |
| social words | 78.43 | 0.32 [0.25; 0.38] | - |
| interrogators | 68.09 | 0.30 [0.23; 0.36] | - |

Topic models

A structural topic modelling approach was used to examine possible changes in topics between phase 1 and 2. This analysis enables us to further examine the content of texts, for example the specific topics that participants worry about [35]. A topic model was constructed for the whole corpus (texts from phase 1 and 2 together), where data collection phase was included as a covariate to assess its effect on topic prevalence. Measures of semantic coherence and exclusivity of topic words [36] were used to determine the ideal number of topics, resulting in a topic model of 15 topics for long texts.

The data collection phase (phase 1 vs phase 2) had a significant effect ($p < .001$) on topic assignment for all topics with the exception of Topic 1 and 4. Table 7 denotes the effect size of the differences in topic proportion between phases. The topics' content demonstrates the differences between the two data collection moments. Several topics contain the word 'vaccine' as a frequent term, all of which significantly increased in the second data collection phase (Topic 14, 7, and 3). Topic 14 and 7, which showed the largest increase in phase 2, seem to refer to hope and a return to normality. In contrast, Topic 8 and 11 largely discuss worries about loved ones - these were significantly more prevalent in the first phase. Negative emotions demonstrated in Topic 10 and 9 were also more prevalent in phase 1. This also holds for topics that seemingly relate to rule following (Topic 6) and panic buying (Topic 13) which were themes that were a lot more prevalent at the start of the pandemic. Interestingly, Topic 4, which appears to refer to a negative outlook on the government handling of the pandemic, did not significantly differ between phase 1 and 2.

Table 7. *Topic terms and differences between phases (long texts)*

| Topic | Cohen's d (99% CI) | Wave 2 vs. wave 1 | Most common terms |
|---|---|---|---|
| 14 | 0.90 [0.83; 0.97] | + | vaccin, back, normal, will, look, year, abl, forward, hope, now, see, feel, famili, school, friend |
| 7 | 0.59 [0.52; 0.66] | + | will, vaccin, hope, get, back, lockdown, think, still, normal, peopl, soon, case, countri, feel, new |
| 8 | 0.60 [0.53; 0.66] | - | worri, famili, also, virus, will, catch, get, concern, children, anxious, time, member, scare, parent, health |
| 3 | 0.57 [0.51; 0.64] | + | vaccin, govern, year, feel, now, like, covid, still, handl, lockdown, peopl, need, way, pandem, seem |
| 11 | 0.42 [0.35; 0.48] | - | work, home, famili, day, friend, stay, get, time, live, miss, see, feel, can, abl, walk |
| 6 | 0.37 [0.30; 0.43] | - | peopl, rule, follow, angri, feel, take, think, everyon, mani, serious, will, get, govern, make, virus |
| 10 | 0.35 [0.29; 0.42] | - | feel, will, worri, anxious, also, know, sad, long, famili, situat, job, health, like, futur, virus |
| 2 | 0.31 [0.25; 0.38] | + | feel, time, thing, life, like, get, normal, just, situat, sad, want, famili, back, miss, see |
| 13 | 0.36 [0.30; 0.43] | - | get, shop, hous, leav, food, virus, just, peopl, news, need, like, now, even, risk, can |
| 9 | 0.26 [0.20; 0.33] | - | futur, will, anxieti, health, fear, anxious, financi, worri, impact, mental, concern, world, may, caus, situat |
| 5 | 0.11 [0.05; 0.17] | - | work, concern, social, distanc, feel, virus, peopl, may, risk, situat, vulner, still, howev, relax, spread |
| 12 | 0.12 [0.05; 0.18] | - | just, much, know, els, want, think, seem, dont, can, anyth, realli, like, say, die, get |
| 15 | 0.12 [0.06; 0.18] | - | situat, feel, Coronavirus, person, time, fear, think, anxious, also, lockdown, make, home, use, difficult, like |
| 1 | 0.07 [0.00; 0.13] | - | now, bring, stori, feel, like, frustrat, apart, control, book, life, green, live, media, bit, orang |
| 4 | 0.03 [-0.03; 0.09] | - | govern, angri, anger, nhs, disgust, also, feel, respons, countri, peopl, situat, staff, mani, handl, lack |

*Note:* all *p* < 0.001 except topics 1 and 4.

*n*-gram differentiation analysis

Next, we investigated which terms shifted the most from the first to the second phase. Specifically, we tested whether the *n*-gram frequencies (unigrams, bigrams and trigrams) differed between the first and the second phase. Since each participant provided two texts, we used a within-subjects test on the *n*-gram frequencies. Since the parametric assumptions are not met for *n*-gram frequencies, we used the Wilcoxon signed rank sum test to test for each *n*-gram whether the frequency between the phases changed. A joint corpus of both phases was created, lower-cased, stemmed, stopwords were removed (using the default English stopword list from the *quanteda* package), and only *n*-grams with a document frequency of at least 5% were retained. The signed rank sum test results in a large proportion of ties with skewed data [37], so we resolved the ties by assigning random ranks to tied cases and ran 500 iterations with random seeds. These were then averaged, and the findings below show the mean and standard error. We use the *r* effect size to show the top 20 most moved *n*-grams in Table 8. Values of *r* closer to +1.00 indicate a higher *n*-gram frequency in phase 2 than in phase 1, and vice versa for values approaching -1.00.

The results suggest that the *n*-grams that increased the most were related to the prospect of the vaccine, the opening of the country and returning back to normal, while decreased terms were about worries, the NHS, staying home, stress, worker and worry - indicating a more worried look on the pandemic in phase 1 than in phase 2.

Table 8. *The top 20 increased and decreased n-grams (ranked by the absolute effect size r)*

| Increased *n*-grams (phase 2 > phase 1) | | Decreased *n*-grams (phase 2 < phase 1) | |
|---|---|---|---|
| *n*-gram | *r* (SE) | *n*-gram | *r* (SE) |
| vaccin | 0.98 (0.00) | worker | -0.67 (0.02) |
| covid | 0.92 (0.01) | stay_home | -0.66 (0.03) |
| look_forward | 0.82 (0.02) | news | -0.62 (0.03) |
| open | 0.80 (0.02) | nhs | -0.57 (0.03) |
| forward | 0.76 (0.02) | stay | -0.56 (0.02) |
| year | 0.69 (0.01) | hous | -0.52 (0.03) |
| school | 0.67 (0.02) | worri | -0.50 (0.01) |
| return | 0.66 (0.02) | serious | -0.50 (0.03) |
| restrict | 0.62 (0.02) | isol | -0.50 (0.03) |
| pandem | 0.62 (0.02) | stress | -0.49 (0.03) |
| handl | 0.60 (0.03) | elder | -0.47 (0.04) |

| first | 0.58 (0.03) | scare | -0.47 (0.02) |
| --- | --- | --- | --- |
| back | 0.58 (0.01) | also_worri | -0.43 (0.03) |
| travel | 0.54 (0.04) | home | -0.43 (0.01) |
| normal | 0.52 (0.01) | tri | -0.42 (0.02) |
| new | 0.50 (0.02) | find | -0.42 (0.03) |
| go_back | 0.50 (0.02) | anxieti | -0.38 (0.02) |
| hope | 0.50 (0.01) | concern | -0.37 (0.02) |
| look | 0.48 (0.02) | die | -0.36 (0.03) |
| wait | 0.48 (0.03) | week | -0.35 (0.03) |

*The effect of the term "vaccine"*

Using the *n*-gram differentiation analysis revealed an important role for the term "vaccine" (or variations thereof such as "vaccinated", "vaccination"). We explored whether references to the vaccine affected the emotion change from phase 1 to phase 2. We counted the occurrences of (versions of) the unigram "vaccine" and then separated the sample into those who did (*n*=1010) and did not (*n*=688) refer to it in the long text of phase 2. Table 9 shows the statistical findings when we assessed the emotion change scores between these two groups. Participants who mentioned the vaccine had a larger reduction in anxiety (*d*=0.22) and sadness (*d*=0.17) between the two phases than those who did not mention it. There were no changes for the other emotions.

Table 9. *Mean (SD) emotion change depending on whether participants mentioned the vaccine in phase 2.*

| **Emotion** | **Vaccine mentioned** | **Vaccine not mentioned** | **Cohen's *d* [99% CI]** | **Bayes Factor** |
| --- | --- | --- | --- | --- |
| *anger* | -0.32 (2.42) | -0.16 (2.63) | 0.06 [-0.07; 0.19] | 0.12 |
| *anxiety* | -1.61 (2.45) | -1.05 (2.57) | 0.22 [0.09; 0.35] | 1241.43 |
| *desire* | 0.66 (2.33) | 0.80 (2.46) | 0.06 [-0.07; 0.19] | 0.12 |
| *disgust* | -0.32 (2.32) | -0.15 (2.57) | 0.07 [-0.06; 0.19] | 0.14 |
| *fear* | -1.95 (2.40) | -1.67 (2.48) | 0.11 [-0.02; 0.24] | 0.74 |
| *happiness* | 0.77 (2.17) | 0.63 (2.33) | 0.06 [-0.07; 0.19] | 0.12 |
| *relaxation* | 0.78 (2.47) | 0.46 (2.54) | 0.13 [0.00; 0.25] | 1.55 |
| *sadness* | -1.06 (2.70) | -0.59 (2.74) | 0.17 [0.05; 0.30] | 26.24 |
| *worry* | -1.69 (1.86) | -1.53 (1.94) | 0.08 [-0.05; 0.21] | 0.18 |

*Note*: Negative *change* values indicate that the value of the emotion went down in phase 2 compared to phase 1. Vice versa for positive values. Values of zero imply no change.

**Discussion**
The COVID-19 pandemic is affecting the lives and mental health of millions. This paper examined how emotional responses to the pandemic changed within one year - from the early onset with severe lockdown measures until the prospect of a return to normal with ongoing vaccination programmes and decreasing incident rates. Using a within-subjects design with a one-year lag, we could assess the change in people's emotional responses to the pandemic. Two core questions guided this paper: (1) How did emotional responses change from April 2020 to April 2021? (2) How can we use text data as a lens through which we can learn about emotional responses in more detail?

*Change in emotions*
The analyses suggest that, on average, the intensity of negative emotions decreased and that of positive emotions increased from the onset of the pandemic to the prospect of re-opening the country. These findings are not surprising: the first phase of data collection happened at a time where unequalled lockdown measures were in place, hence resulting in high scores for negative emotions and low scores for positive emotions. These then moved towards the positive in phase 2. The correlations between the two data collection points suggest that it is unlikely that the change towards less extreme scores in phase 2 is attributable to a regression to the mean. Instead, we found subgroups hinting at a more complex picture.

Two clusters of participants emerged: one group exhibited a pattern in emotion change that we identified as *well-coping*. Participants in that group showed a general increase in positive emotions and a decrease in worry-related emotions. In contrast, a second group appeared to be less well-adapted after one year in the pandemic, with a pattern that we interpret as resembling *resignation*. Importantly, we failed to find straightforward explanations for these cluster patterns. While participants in the well-coping cluster were marginally older than those in the resignation cluster, neither gender nor linguistic differences emerged. Previous work [6] found gender differences in the topics of the first phase texts, but these patterns did not seem to be at play with the subgroups. Possibly, membership in either of the two clusters can be inferred from the texts that participants wrote. If that were the case, we would expect at least some predictive relationship between linguistic features of the text data and the participants' cluster membership. We tested this with supervised machine learning but failed to find an above-guessing prediction performance (see Appendix).

*Learning about emotions through text data*
Linguistic analyses revealed marked differences between data collection phases. For instance, using LIWC variables, we found increased positive emotion and decreased negative emotion in phase 2 compared to phase 1. References to the home also decreased in phase 2, possibly reflecting the difference in social distancing rules between phase 1 (strict lockdown at home) and 2 (easing of social distancing rules).

When we applied structural topic modelling, we observed dominant themes for each phase (April 2020 vs April 2021) in participants' writing. Similar to the LIWC analysis, we observed more reference to negative emotion about the pandemic in phase 1 than in phase 2. We also shed some light on the content of worries, which often related to family and friends. In the second phase, topics relating to (hope for) a return to normality often included references to the vaccine. One of the topics which discussed a negative outlook on the government's handling of the pandemic did not differ between data collection phases. These results possibly suggest that UK residents were dissatisfied with government handling during both phases of data collection but had different concerns pertaining to their individual situation (worries in phase 1 and hope in phase 2). Although several topics showed overlap in most frequent terms and thus were not easily distinguishable, topic modelling enabled us to shed some further light on the content of worries and hope demonstrated by participants in the study.

Further trends were revealed in the *n*-gram differentiation analysis. The frequencies of *n*-grams related to the vaccine and an outlook on the future ("looking forward to") were markedly higher in the second phase than in the first phase. Conversely, the mentions of *n*-grams, including "worry", were more frequent in the first phase. These findings highlight that text data somewhat capture significant changes in society and that participants have them on their minds. The vaccine was the dominant word in phase 2, which mirrors the public discourse in the UK of the time of data collection. Interestingly, the notion of a vaccine was practically absent in the first phase. Follow-up analyses showed that individuals who mentioned the vaccine showed a more pronounced reduction in anxiety and sadness than those who did not refer to the vaccine.

*So what? The bigger picture and an outlook*

A key challenge for public health may lie in the mental health effects that the COVID-19 pandemic is having. With limited resources available, research such as the current study can help health care professionals identify individuals who may be at a heightened risk of more severe mental health problems. Related work, for example, has looked at changes in loneliness due to the various interventions introduced due to the pandemic [38,39]. Understanding the relationship between an emotional response such as loneliness and text data produced by people is a promising avenue for research but is still in an early stage [40]. Importantly, our analysis showed that it is oversimplified to assume that a population processes the pandemic in a homogenous manner. Two radically divergent subgroups were identified. This, again, may aid mental health professionals and policymakers in their approach to handling the (mental health) consequences of the pandemic. Future work could test *i)* how subgroups can be identified and *ii)* whether membership in a subgroup also corresponds to differences in mental health and possibly even the effectiveness of different intervention strategies.

Our paper makes the connection between emotion data and text data. Although similar connections have been made in the past (e.g., [19,20] they lacked ground truth of the emotional states of text writers. By assessing 'ground truth' emotional states (albeit self-reported ones) and having participants write about them, we were able to examine linguistic correlates of emotions and worries. For clarification, the term 'ground truth' emotional states here denotes that the emotional states of individual participants were provided by the participants themselves rather than through an external annotation procedure. While we appreciate that emotional states are volatile, our dataset records the participants' emotional wellbeing at the time of conducting the task, thus representing their ground truth mental states when writing the statements. All in all, this approach can help us as a research community to test and ultimately harness the simmering potential of observational text data. Future work could, for example, use the dataset to assess whether text data from one phase can be used to predict emotions in a subsequent phase. Ideally, social and behavioural science research continues to adopt methods from the computational sciences and builds on the current work to further our understanding of text data as a means to study behaviour and emotions.

*Limitations*

Our paper comes with a few limitations worth noting. First, we only captured the emotional responses of UK residents. While this was done to avoid the influence of potential confounds such as government responses, this somewhat limits the ecological validity of these findings. Whether our results are valid in other countries is an open question and hard to ascertain absent longitudinal data on emotion and text data from other countries. Second, as with all *unsupervised* techniques, the topic models and the clustering analysis contain subjectivity in the interpretation. Other researchers may have interpreted the clusters differently and hence yielded different conclusions. We acknowledge that subjectivity, and by making the raw data available, we encourage others to challenge our findings. Similarly, we have only focused on the core of the dataset but have not examined further variables such as socioeconomic status or self-assessed linguistic expression ability. All these data are now publicly available as part of this paper. Third, we have identified the two subgroups only through exploratory analyses and refrained from conducting further in-depth analyses on the properties of these groups. Ideally, similar studies could attempt to replicate the findings and solidify the evidence favouring heterogenous coping with the pandemic. Future work could then look at linguistic differences between these clusters (i.e., do the subgroups use different words and topics?).

**Conclusion**

The COVID-19 pandemic could have potentially long-lasting consequences on public mental health. Understanding the emotional responses to a "new normal" requires an analytical toolkit that allows for the complexity in how people (struggle to) cope. This paper identified a heterogeneous pattern in people (well-coping vs resigned ones). Our method demonstrated how text data can help us understand the impact of the COVID-19 pandemic on people's lives, and how individuals work through and adapt to drastically changing circumstances. We believe that our data collection efforts and subsequent analyses represent a valuable resource to better understand emotional standpoints of the public, and we encourage other researchers to further exploit the potential of our presented dataset.

**Acknowledgment**

We thank two anonymous reviewers for suggestions that helped us improve the paper; in particular the exploration of the role of the term "vaccine" on emotion change scores.